\documentclass[10pt,twocolumn,letterpaper]{article}

\usepackage{cvm}
\usepackage{times}
\usepackage{epsfig}
\usepackage{graphicx}
\usepackage{amsmath}
\usepackage{amssymb}
\usepackage{eso-pic}

\usepackage{booktabs}
\usepackage{multirow}
\usepackage{xcolor}    
\usepackage{soul}      

\sethlcolor{lightgray} 
\usepackage[pagebackref=true,breaklinks=true,letterpaper=true,colorlinks,bookmarks=false]{hyperref}

\newcommand{\keywords}[1]{{\bf \emph{Keywords: #1}}}
\cvmfinalcopy

\def\cvmPaperID{511} 

\ifcvmfinal\fi
\begin{document}

\title{MASS: Mesh-inellipse Aligned Deformable Surfel Splatting for Hand Reconstruction and Rendering from Egocentric Monocular Video}

\author{Haoyu Zhu \and Yi Zhang \and Lei Yao \and Lap-pui Chau \and Yi Wang\\
Department of Electrical and Electronic Engineering \\
The Hong Kong Polytechnic University, Hong Kong \\
{\tt\small yi-eie.wang@polyu.edu.hk}
}

\maketitle

\begin{abstract}
    Reconstructing high-fidelity 3D hands from egocentric monocular videos remains a challenge due to the limitations in capturing high-resolution geometry, hand-object interactions, and complex objects on hands. Additionally, existing methods often incur high computational costs, making them impractical for real-time applications. In this work, we propose \underline{M}esh-inellipse \underline{A}ligned deformable \underline{S}urfel \underline{S}platting (MASS) to address these challenges by leveraging a deformable 2D Gaussian Surfel representation. We introduce the mesh-aligned Steiner Inellipse and fractal densification for mesh-to-surfel conversion that initiates high-resolution 2D Gaussian surfels from coarse parametric hand meshes, providing surface representation with photorealistic rendering potential. Second, we propose Gaussian Surfel Deformation, which enables efficient modeling of hand deformations and personalized features by predicting residual updates to surfel attributes and introducing an opacity mask to refine geometry and texture without adaptive density control. In addition, we propose a two-stage training strategy and a novel binding loss to improve the optimization robustness and reconstruction quality. Extensive experiments on the ARCTIC dataset, the Hand Appearance dataset, and the Interhand2.6M dataset demonstrate that our model achieves superior reconstruction performance compared to state-of-the-art methods.
\end{abstract}

\section{Introduction}
Reconstructing and rendering high-fidelity 3D hand models from egocentric RGB videos enables the generation of diverse, high-quality visual displays for tasks such as rendering and synthetic datasets generation. This task has wide-ranging applications, including virtual reality (VR) \cite{Liu2024TouchscreenbasedHT} and augmented reality (AR) \cite{Tang_2021_ICCV}. Personalized 3D hand rendering plays a key role in data augmentation for gesture recognition and robotic manipulation \cite{qu2025hogsabimanualhandobjectinteraction}. Yet, it remains exceptionally challenging due to self-occlusions, rapid motion, limited viewpoints, and complex hand–object interactions.

\begin{figure}[t]
  \centering
  \includegraphics[width=\columnwidth]{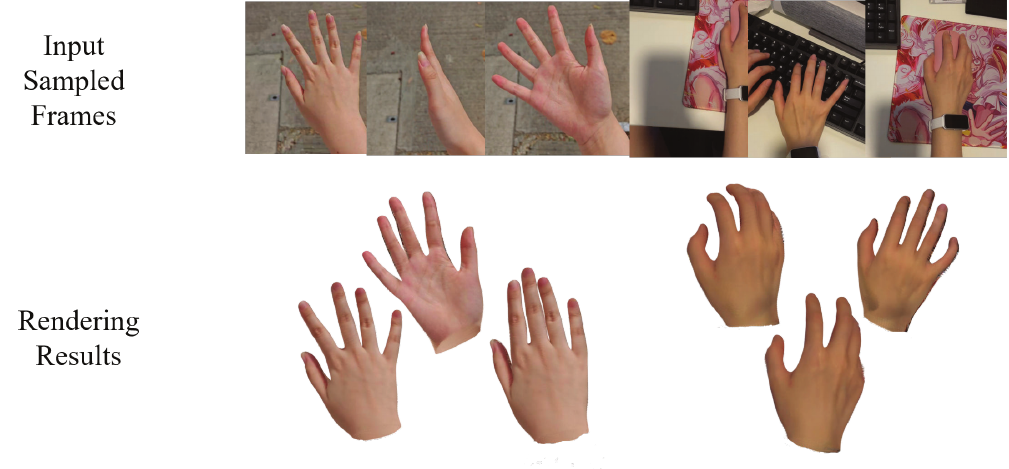}
  \caption{Hand reconstruction on real-world phone-shot videos. Extracted frames from two videos are shown in the first row. The second row images depict renderings of reconstructed hands.}
  \label{fig:teaser}
\end{figure}

Existing methods for 3D hand reconstruction fall into two camps. The implicit and explicit representations. Implicit methods, such as Neural Radiance Fields (NeRF)-based methods \cite{10611230,9878817,10377744} reconstruct the geometry and appearance with neural representations. While implicit representations like NeRF achieve high visual fidelity, they suffer from slow rendering and high memory costs. On the other hand, explicit parametric models such as MANO \cite{10.1145/3130800.3130883}, and the extensions like HTML \cite{10.1007/978-3-030-58621-8_4} and HARP \cite{karunratanakul2023harp}, rely on parametric hand meshes to model hands with appearance. They are efficient but lack the expressivity to capture fine-grained deformations or hand–object contact details. Recent advances in 3D Gaussian Splatting\cite{10.1145/3592433} offer real-time rendering, yet they struggle with monocular egocentric settings due to under-constrained geometry and poor surface alignment. 

To bridge this gap, we propose Mesh-inellipse Aligned deformable Surfel Splatting (MASS), a novel framework that combines the structural prior of parametric hand models with the rendering fidelity of 2D Gaussian surfels. It is designed to reconstruct a precise hand from egocentric monocular RGB video. Our method leverages a mesh-aligned Steiner Inellipse initialization to ensure geometric fidelity, introduces a neural deformation module for dynamic detail, and employs a two-stage optimization strategy for stable training. As shown in Fig.\ref{fig:teaser}, MASS reconstructs high-fidelity hands from casually captured phone videos. MASS accurately captures skin texture.

MASS consists of two key components, each addressing specific challenges in 3D hand reconstruction. In order to achieve the end-to-end high-fidelity geometry and texture learning, we propose the \textit{first} module, Mesh-to-Surfel Conversion. The module directly converts the mesh template into a high-resolution 2D Gaussian Surfel representation \cite{2dgs} without adaptive density control in 3DGS \cite{10.1145/3592433}. 2D Gaussian surfels explicitly align intricate planar surfaces, reduce artifacts commonly observed in 3D Gaussian with less constrained data, and provide consistent rendering values across viewpoints. This direct conversion is guided by the Steiner Inellipse of each triangular mesh face, which allows us to compute surfel attributes such as centroid, scale, and rotation with high precision. Unlike previous works \cite{shao2024splattingavatar} that treat Gaussian initiation as a random process or simply placing on mesh vertices, our method avoids convergence and instability issues by leveraging the mesh structure for initialization, ensuring that the surfels registration aligns tightly with the underlying geometry. 
This module effectively bridges the gap between coarse parametric meshes and high-fidelity Gaussian splatting representations. 

The \textit{second} module, Gaussian Surfel Deformation, addresses the challenge of modeling flexible deformations and dynamic hand motion. Hand-object interactions in egocentric videos often involve complex movements that cannot be captured by linear blend skinning. To solve this, we design a neural deformation network that predicts residual updates for surfel geometry with a learnable opacity to adaptively adjust surfels. With a multi-resolution hashgrid encoder, a neural feature encoder, and a decoder, the deformation network enables accurate modeling of complex deformations and unregistered regions of the hand. 

To achieve photorealistic rendering with egocentric reconstruction. We adaptively design the optimization process to address the impact of inaccurate camera pose estimation and the instability of training of the flexible deformations module. We propose a two-stage training strategy in order to decouple geometry and texture optimization with image silhouette and geometry supervision. In the first stage, geometry attributes and low-frequency harmonics coefficients are optimized with attached Gaussian surfels to avoid instability during the early stage of training. In the second stage, opacity masks and high-frequency harmonics coefficients are refined for detailed texture representation. Additionally, we introduce a novel binding loss that ensures training robustness and consistency between deformed surfels and the original mesh geometry in the early training stage.

Extensive experiments were conducted on ARCTIC \cite{fan2023arctic} and Hand Appearance \cite{karunratanakul2023harp}. Our method achieves faster rendering speed and provides high-fidelity rendering results (see Fig.\ref{fig:teaser}), outperforming state-of-the-art methods, such as HARP \cite{karunratanakul2023harp} and 3D-PSHR \cite{JIANG2025111426}. 
The contributions of our paper are summarized as follows:

\begin{itemize}
    \item We propose a novel Mesh-to-Surfel Conversion module that effectively solves conversion between parametric meshes directly and 2D Gaussian surfels with a designed structure, enabling high-resolution geometry and texture representation.  
    \item  We introduce a Gaussian Surfel Deformation module that models dynamic, enhancing the flexibility of hand deformations using a neural deformation network, allowing for precise adaptation to complex motions.
    \item We develop a two-stage training strategy and a novel binding loss to optimize surfel attributes efficiently while ensuring geometric consistency. 
    \item Our method achieves state-of-the-art performance on egocentric hand reconstruction and rendering, outperforming existing approaches in both accuracy and efficiency.
\end{itemize} 

\section{Related Works}
3D hand reconstruction from monocular video has seen significant progress through both explicit mesh-based and implicit neural representations, each offering distinct trade-offs in fidelity, efficiency, and expressiveness.

\subsection{Explicit Hand Reconstruction}
Mesh-based methods have been widely adopted due to their compatibility with established computer graphics and visual effects pipelines. Early works such as MANO \cite{10.1145/3130800.3130883} and SMPLX \cite{8953319} introduced parametric models for the human hand and body meshes, incorporating robust shape and pose priors to effectively capture hand articulation and body motion. Subsequent works such as HTML \cite{10.1007/978-3-030-58621-8_4}, NIMBLE \cite{10.1145/3528223.3530079}, and HARP \cite{karunratanakul2023harp} extended MANO by modeling non-rigid deformations and exploring texture feature learning. However, the reliance on template models constrains their expressivity to represent complex geometry and appearance with motion variations.

Recently, 3D Gaussian Splatting (3DGS) \cite{10.1145/3592433} has emerged as a promising representation for hand reconstruction and rendering. Unlike NeRF, 3DGS offers high fidelity, real-time rendering, and reduced memory requirements, making it suitable for dynamic and interactive applications. Methods like MANUS \cite{10658236} extend the 3DGS framework to achieve efficient and high-fidelity hand reconstruction by introducing an articulated 3D Gaussian representation for markerless capture of human hand grasps. However, MANUS requires multi-view setups for optimal performance and has limited applicability in egocentric monocular scenarios. In contrast, surface representations such as 2D Gaussian surfels can more accurately fit intricate surfaces while reducing noise and artifacts commonly observed in overlapping 3D Gaussian spheres.  3GS-based  hand reconstruction methods, like 3D-PSHR \cite{JIANG2025111426} employs a framework with dynamic upsampling and deformation, 3D-PSHR achieves real-time performance and robust geometry fitting. While 3D-PSHR supports both single-view and multi-view setups, its reliance on explicit point clouds can result in a lack of fine detail in the reconstructed texture, leading to inaccuracies in texture representation. Additionally, point-based splatting methods may struggle to achieve the same level of realism as implicit approaches when modeling complex hand-object interactions or fine-grained appearance changes.

\subsection{Implicit Hand Reconstruction}
Implicit neural representations, such as Neural Radiance Fields (NeRF), have been extensively applied to hand reconstruction tasks due to their ability to model complex shapes and appearances. However, it struggles with limited accuracy and requires significant computational resources, including high memory usage and significant training and inference costs. The LISA framework \cite{9878817} pioneered a neural approach to modeling human hands by disentangling shape, pose, and color representations. LiveHand \cite{10377744} enables real-time rendering of photorealistic hands but relies heavily on pre-training with large-scale multi-view datasets. HandNeRF \cite{10611230} introduces a method to jointly reconstruct the geometries of hands and objects from a single image by encoding correlations between 3D hand features and 2D object features. 

In addition to NeRF-based methods and Gaussian-based methods, other implicit representation approaches have been proposed. For example, OHTA \cite{10658610} introduces a one-shot framework for animatable hand avatar creation using a single RGB image. It employs a Hand Prior Network (HPNet) to encode geometry, texture, and shadow priors, enabling high-fidelity hand reconstructions. However, OHTA is limited to static single-image inputs and does not handle dynamic sequences or temporal consistency. 


\subsection{Dynamic Hand Modeling from Egocentric Video}
With improvements driven by scaling up both the training data and model capacity, recent advances have leveraged parametric model regression combined with large-scale pre-trained Transformer architectures to set new standards of hand pose reconstruction, predicting camera and pose parameters as tokens. For example, HAMER \cite{10655481} reconstructs 3D hand meshes from monocular images captured in either third-person or egocentric views. To address temporarily consistent hand reconstruction, numerous works have further built upon HAMER to enhance performance in specific applications. Extensively, WiLoR \cite{potamias2025wilorendtoend3dhand} tackles challenges related to multi-target scenarios and real-time requirements, while Dyn-HaMR \cite{yu2024dynhamrrecovering4dinteracting} and HaPTIC \cite{ye2025predicting4dhandtrajectory} focus on ensuring global pose consistency and the continuity of hand motion in continuous estimation. Our method is designed for monocular videos, built on a Transformer-based reconstruction to enable end-to-end hand geometry and texture learning. 

However, existing approaches struggle to simultaneously achieve high geometric fidelity, efficient dynamic modeling, and robust training in hand reconstruction and rendering under egocentric monocular constraints—limitations that our Mesh-in-ellipse Aligned deformable Surfel Splatting (MASS) framework is designed to overcome.

\begin{figure*}[t]
  \centering
  \includegraphics[width=0.95\linewidth]{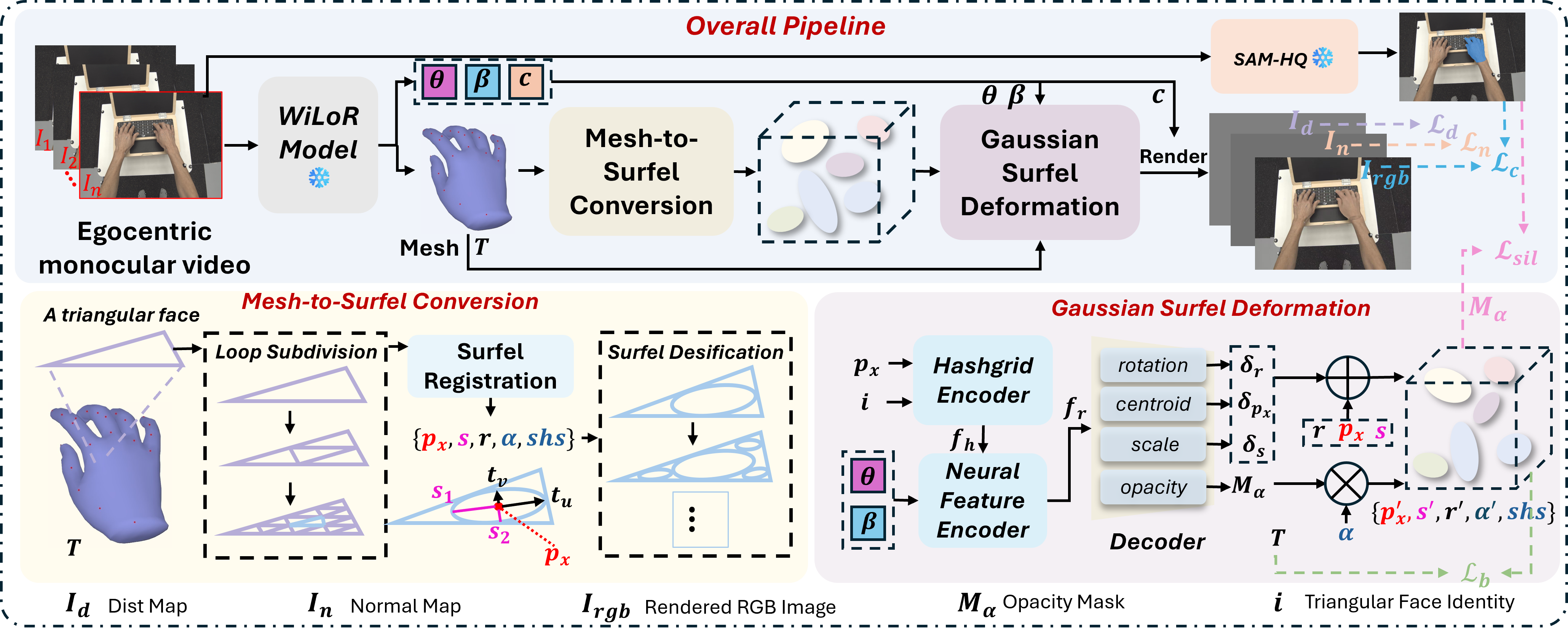}
  \caption{Overall pipeline of our method. Starting with an egocentric monocular RGB video$\{{I_k}\}_{k=1}^{n}$, our pipeline first generates a parametric hand mesh $\boldsymbol{T}$ using a pre-trained model. The Mesh-to-Surfel Conversion module then transforms the coarse mesh $\boldsymbol{T}$ into a high-resolution 2D Gaussian Surfel representation, initializing surfel attributes (centroid $\boldsymbol{p_{x}}$, scale $\boldsymbol{s}$, and rotation $\boldsymbol{r}$). Next, the Gaussian Surfel Deformation module refines these attributes by predicting residual updates ($\boldsymbol{\delta p_{x}}$, $\boldsymbol{\delta s}$, $\boldsymbol{\delta r}$) and generating an opacity mask $\boldsymbol{M_{\alpha}}$ to model complex hand deformations. Finally, the refined surfels are rendered following the 2D Gaussian Splatting pipeline, producing high-fidelity hand geometry and texture reconstructions. The optimization losses include image-based reconstruction loss $\mathcal{L}_{c}$, geometry constraints $\mathcal{L}_{d}$, and $\mathcal{L}_{n}$, and weak geometry supervision $\mathcal{L}_{sil}$ from the SAM-HQ segmentation result.}
  \label{fig:pipeline}
\end{figure*}
\section{Methodology}

\subsection{Overall Pipeline}
As shown in Fig.\ref{fig:pipeline}, the proposed pipeline reconstructs precise hand geometry with texture features from egocentric monocular RGB video. To achieve this, we introduce two key modules: Mesh-to-Surfel Conversion and Gaussian Surfel Deformation. The pipeline consists of four main stages: Preprocessing, Mesh-to-Surfel Conversion, Gaussian Surfel Deformation, and Rendering and Optimization.

\subsection{Mesh-to-Surfel Conversion.}

To achieve the photo-realistic rendering only using the prior of the MANO parametric model, we propose a Mesh-to-Surfel Conversion module that transforms the coarse MANO mesh into a high-resolution 2D Gaussian Surfel (2DGS) representation. 2D Gaussian surfels \cite{2dgs} represent surfaces as oriented elliptical Gaussian disks parameterized by the centroid $\boldsymbol{p_x}$, scale $\boldsymbol{s}$, and rotation $\boldsymbol{r}$, along with opacity $\boldsymbol{\alpha}$ and view-dependent appearance attributes $\boldsymbol{shs}$. 2D Gaussian surfels provide an intermediate representation that connects 3DGS and surface geometry. 3DGS can be transferred from surfel representation by setting the third component of the 3DGS's scale to near zero along the normal vector \cite{ye2024gaustudio,wen2024gomavatar}. This enables our work to be integrated into other 3DGS-based and mesh-based workflows.

We initialize the geometric surfel attributes (centroid, scale, and rotation) directly from the deformed mesh template using the properties of the Steiner Inellipse, ensuring that the attached surfels are bonded to the correct surface geometry. Unlike $SE(3)$ or affine deformation fields that operate in 3D space and require careful regularization to avoid self-intersections or drift, surfel-based splatting operates in a 2.5D local parameterization, naturally constraining deformations to plausible surface offsets. This design aligns with the observation that fine hand details predominantly manifest as normal-direction displacements rather than full 3D warps.
Providing bonded surface avoids the convergence and instability issues observed when only initiating detached Gaussian surfels. Besides, mesh and surfel subdivision enable high-fidelity rendering. The conversion consists of two main steps: mesh refinement and surfel registration, as described in detail below.

\textbf{Mesh refinement.} The process begins with refining MANO model \cite{10.1145/3130800.3130883} mesh template with shape and pose parameters.  The shape parameters $\boldsymbol{\beta}$ describe the geometry feature of the hand using coefficients of a PCA-based model. The pose parameters $\boldsymbol{\theta}$ define the rotation angles between adjacent joints. The MANO mesh $\boldsymbol{T}$ contains the 3D coordinates of all vertices ${v_k}$ and the mesh face identities ${i}$, where $v_k$ represents the $k$-th vertex in the mesh and $i$ represents the $i$-th triangular face in the mesh. 

To prepare the mesh for surfel registration, we refine its resolution using loop subdivision \cite{Loop1987SmoothSS}, which subdivides each triangular face of the mesh into smaller triangles. The loop subdivision provides a relatively average densification effect across the mesh. To avoid over-smoothing and a plain visual appearance—particularly in hand rendering due to scale changes, we introduce Fractal Densification.
See from the surfel densification part of Fig.\ref{fig:pipeline}, Steiner Inellipse covers most of the original triangle. However, in our hypothesis, the remaining area close to the vertices is supposed to be the area of high-frequency detail. We were inspired by the Sierpinski triangle, using Fractal Densification to fit in the remaining area.

\textbf{Surfel registration.} After refinement, the module registers attached 2D Gaussian surfels for each triangular subdivided mesh face. A surfel compactly represents local geometry and texture. The geometric attributes of each surfel, including its centroid $\boldsymbol{p_x}$, scale $\boldsymbol{s}$, and rotation $\boldsymbol{r}$, are directly derived from the geometry of the underlying triangular face using the properties of Steiner Inellipse.  In our proposed method, the opacity $\boldsymbol{\alpha}$ does not involve the density control. The opacity and appearance $\boldsymbol{shs}$ are view-dependent as learnable parameters for rendering and optimization for flexible surface displacement.

In the surfel registration process.  The centroid $\boldsymbol{p_x}$ of each surfel is computed as the average of the three vertices of the triangular face:
\begin{equation}
    \boldsymbol{p_x} = \frac{1}{3}(\boldsymbol{v_A} + \boldsymbol{v_B} + \boldsymbol{v_C}),
\end{equation}
where $\boldsymbol{v_A}$, $\boldsymbol{v_B}$, and $\boldsymbol{v_C}$ are the 3D coordinates of the triangle’s vertices. This centroid serves as the central position of the surfel and ensures that the surfel is positioned accurately within the hand geometry.

The scale $\boldsymbol{s}$ of the surfel is derived from the Steiner Inellipse, which is the largest inscribed ellipse that fits inside the triangle. The ellipse is tangent to the midpoints of the triangle's edges and provides a robust geometric basis for estimating the surfel's size. The edge lengths of the triangle, denoted as $a, b, c$, are used to compute scale $\boldsymbol{s} = [s_1, s_2]$ of the ellipse:
\begin{equation}
\begin{array}{l}
     s_{1} = \frac{1}{6} (a^{2}+b^{2}+c^{2} + 2 F)^\frac{1}{2}\\ s_{2} = \frac{1}{6} (a^{2}+b^{2}+c^{2} - 2 F)^\frac{1}{2}, 
\end{array}
\end{equation}
where F is given by: 
\begin{equation}
F=(a^{4}+b^{4}+c^{4}-a^{2} b^{2}-b^{2} c^{2}-c^{2} a^{2})^\frac{1}{2} .
\end{equation}

The rotation vector $\boldsymbol{r} = (\boldsymbol{t_u}, \boldsymbol{t_v})$ defines the orientation of the surfel within the tangent plane of the triangular face, where $\boldsymbol{t_u}$ and $\boldsymbol{t_v}$ are orthogonal basis vectors aligned with the major and minor axes of the Steiner Inellipse. These vectors are computed as follows.
The vertices of the triangle are first represented in the tangent plane using Euler’s formula:
\begin{equation}
    A = r_Ae^{i\theta_A}, \text{ } B = r_Be^{i\theta_B}, \text{ } C = r_Ce^{i\theta_C},
\end{equation}
where $r_A$, $r_B$, $r_C$ are the distances of the vertices from the centroid, and $\theta_A$, $\theta_B$, $\theta_C$ are the angular positions of the vertices relative to the positive $u$-axis.

The focus $Z$ of the Steiner Inellipse, which helps determine the orientation and alignment of the ellipse within the tangent plane, is computed as:
\begin{equation}
Z =\frac{1}{3}(A+B+C \pm (A^{2}+B^{2}+C^{2}-B C-C A-A B)^\frac{1}{2}) 
\end{equation}
Then, the 3D vertices $\boldsymbol{v_A}$, $\boldsymbol{v_B}$, $\boldsymbol{v_C}$ are projected onto the triangle's tangent plane to obtain their Cartesian coordinates $(x, y)$: 
\begin{equation}
    \boldsymbol{v_A} \rightarrow (x_A, y_A), \text{ } \boldsymbol{v_B} \rightarrow (x_B, y_B), \text{ } \boldsymbol{v_C} \rightarrow (x_C, y_C). 
\end{equation}
Using the projected Cartesian coordinates, the covariance matrix $\boldsymbol{C}$ of the triangle can be computed. The eigenvectors of $C$ represent the directions of the major and minor axes of the Steiner Inellipse. These eigenvectors are used to define the rotation vector:
\begin{equation}
    \boldsymbol{r} = (\boldsymbol{t_u}, \boldsymbol{t_v}) = \text{Eig}(\boldsymbol{C}). 
\end{equation}

\subsection{Gaussian Surfel Deformation}
As the original mesh surface and its attached surfels are limited to template geometry, we introduce a Gaussian Surfel Deformation module that predicts the residual updates for the attributes of detached surfels using a neural network-based approach.

\begin{table}
\centering
 \begin{tabular}{ccccc}
 \toprule
  Test-set & Methods & LPIPS $\downarrow$ & PSNR $\uparrow$ & MS-SSIM $\uparrow$ \\ 
 \midrule
  \multirow{2}{*}{waffle}& HARP& 0.0345 & 28.74 & 0.9728 \\
  &Ours&  \textbf{0.0328} & \textbf{34.33} & \textbf{0.9783} \\ \hline
 \multirow{2}{*}{capsule}& HARP& \textbf{0.0359} & 29.37 & 0.9779 \\
 &Ours&  0.0381 & \textbf{31.22} & \textbf{0.9816}\\ \hline
 \multirow{2}{*}{phone}& HARP& 0.0317 & 30.18 & 0.9767 \\
 &Ours&  \textbf{0.2740} & \textbf{32.50} & \textbf{0.9807}\\ \hline
 \multirow{2}{*}{notebook}& HARP& 0.0342 & 29.02 & 0.9755 \\
 &Ours& \textbf{0.0271} & \textbf{38.63} & \textbf{0.9833}\\ \hline
 \multirow{2}{*}{scissors}& HARP& 0.0232 & 28.67 & 0.9804 \\
 &Ours& \textbf{ 0.0141} & \textbf{38.81} & \textbf{0.9929}\\
 \bottomrule
 \end{tabular}
\caption{Quantitative Evaluation on the ARCTIC dataset \cite{karunratanakul2023harp}.}
\label{tab:arctic_comps}
\end{table}

\begin{figure}[t]
  \centering
  \includegraphics[width=\columnwidth]{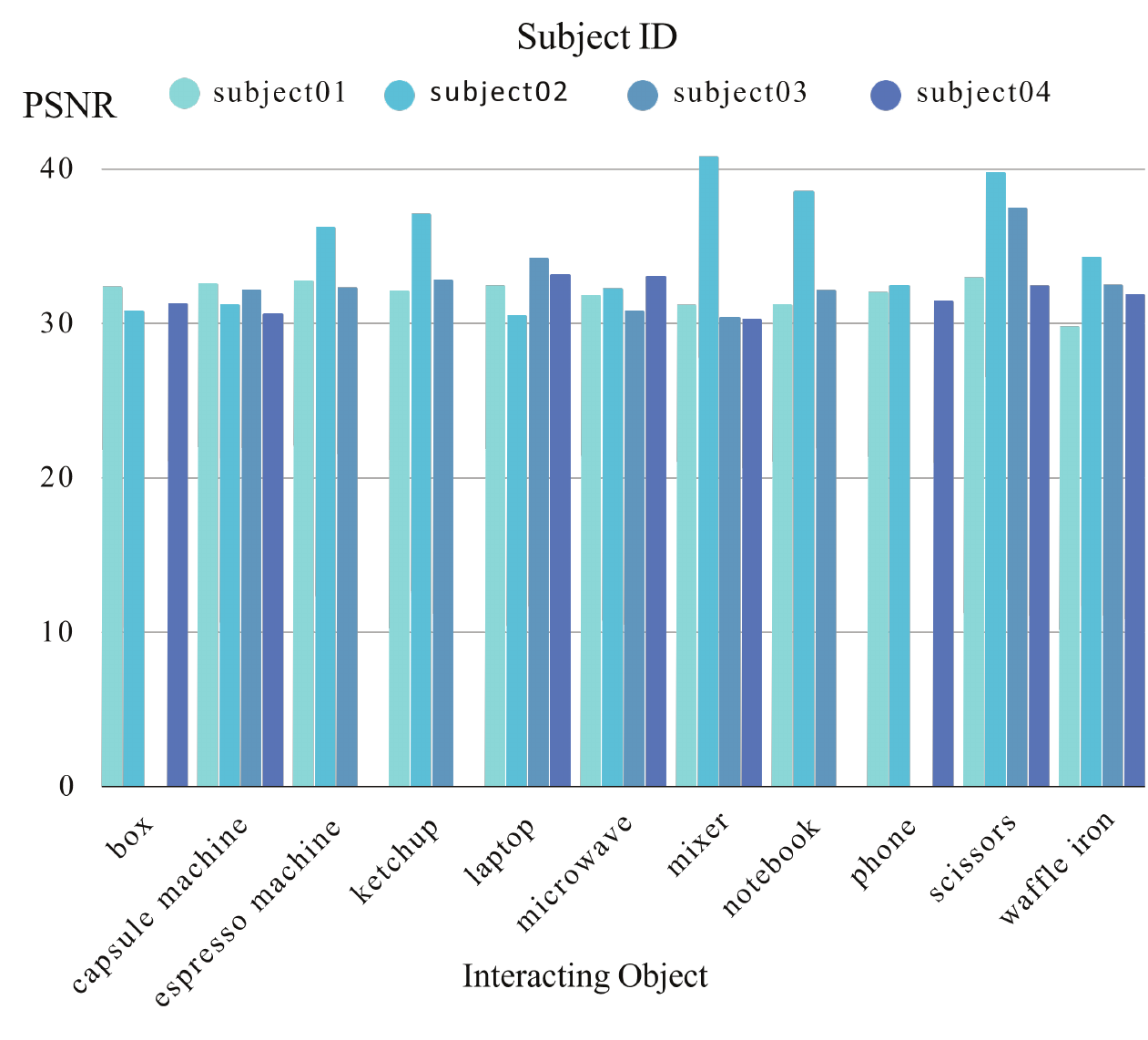}
  \caption{Varied Interacting objects and subjects of the quantitative evaluation on the ARCTIC dataset \cite{fan2023arctic}. The Y-axis represents the best PSNR evaluation result on each test set.}
  \label{fig:histogram}
\end{figure}

\begin{figure}[ht]
  \centering
  \includegraphics[width=\columnwidth]{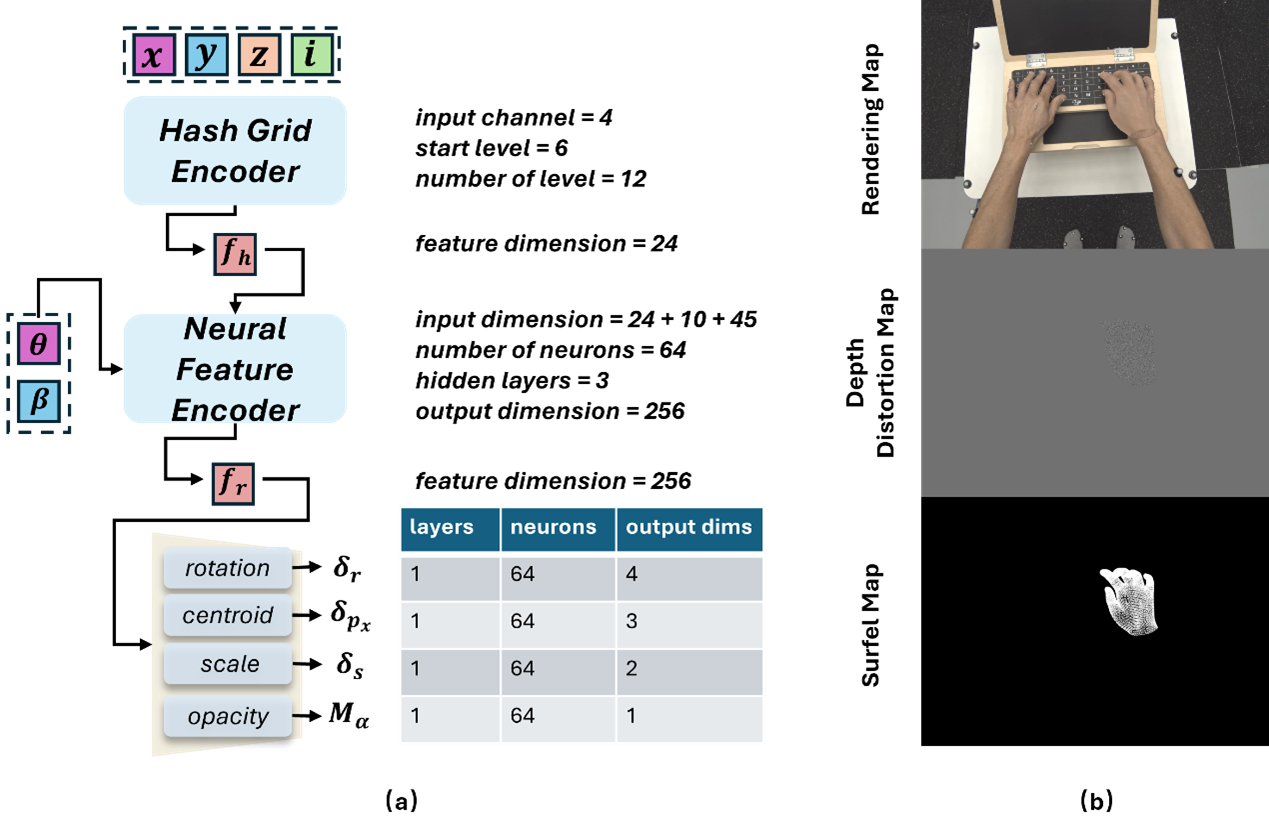}
  \caption{(a) represents the architecture of the deformation module; (b) demonstrates some intermediate rendering results. Depth Distortion Map shows the inconsistency of the intersecting surfels. Surfel Map indicates the center of surfels with white small disks.}
  \label{fig:deform}
\end{figure}

As shown in Fig.\ref{fig:deform} (a), the deformation of the converted surfels is modeled using three networks: a multi-resolution hashgrid encoder, a neural feature encoder, and a decoder. These networks estimate the residual attributes, including the centroid position, scale, and rotation. Below, we describe the role of each network in detail.

\textbf{Multi-resolution hashgrid encoder. }The first component is a multi-resolution hashgrid encoder $\text{Enc}_h$ proposed in Instant-NGP \cite{10.1145/3528223.3530127}. The inputs to the hashgrid encoder include the 3D coordinates of the Gaussian surfel centroid $\boldsymbol{p_x}$ and the initial mesh face identity $i$. The identity $i$ indicates the triangular face to which the surfel belongs. The hashgrid encoder outputs a feature vector $\boldsymbol{f_h}$, which captures localized deformation-relevant information:
\begin{equation}
    \boldsymbol{f_h} = \text{Enc}_h ([\boldsymbol{p_x}, i]).
\end{equation}

\textbf{Neural feature encoder. }The second component is a neural feature encoder $\text{Enc}_n$ composed of fully connected layers. This network refines the features extracted by the hashgrid encoder $\boldsymbol{f_h}$ and incorporates global deformation context by concatenating $\boldsymbol{f_h}$ with the shape parameters $\boldsymbol{\beta}$ and pose parameters $\boldsymbol{\theta}$ obtained from the preprocessing stage. The neural feature encoder processes this concatenated input through a series of fully connected layers to produce a refined feature representation $\boldsymbol{f_r}$:
\begin{equation}
    \boldsymbol{f_r} = \text{Enc}_n([\boldsymbol{f_h}, \boldsymbol{\theta}, \boldsymbol{\beta}]).
\end{equation}

\textbf{Decoder. }The final component is a decoder $\text{Dec}$ which predicts the residual updates for the surfel attributes. Specifically, the decoder outputs the residuals for the centroid $\boldsymbol{\delta p_{x}}$, scale $\boldsymbol{\delta s}$, and rotation $\boldsymbol{\delta r}$ of each surfel. Besides, it also generates an opacity mask $\boldsymbol{M_{\alpha}}$ which adaptively adjusts the opacities $\boldsymbol{\alpha}$ of all surfels. The decoder takes the refined feature representation $\boldsymbol{f_r}$ as input: 
\begin{equation}
    [\boldsymbol{\delta p_{x}},\boldsymbol{\delta s},\boldsymbol{\delta r}, \boldsymbol{M_{\alpha}}] = \text{Dec}(\boldsymbol{f_r}).
\end{equation}

The predicted residuals are used to refine the initial surfel attributes. The final deformed attributes are computed as:
\begin{equation}
    \boldsymbol{p_x^{'}} = \boldsymbol{p_x} + \boldsymbol{\delta p_x}, \text{ } \boldsymbol{r^{'}} = \boldsymbol{r} + \boldsymbol{\delta r}, \text{ } \boldsymbol{s^{'}} = \boldsymbol{s} + \boldsymbol{\delta s}, \text{ } \boldsymbol{\alpha^{'}} = \boldsymbol{\alpha} \otimes \boldsymbol{M_{\alpha}}
\end{equation}

We hypothesize that the combination of a multi-resolution hashgrid encoder and a lightweight MLP strikes an optimal balance between local geometric expressivity and global deformation consistency: the hashgrid efficiently captures high-frequency, spatially localized hand deformations, while the MLP integrates global pose and shape priors to ensure coherent, physically reasonable motion across the entire hand surface—all while maintaining computational efficiency due to the sparse, hierarchical nature of the hashgrid and the compactness of the MLP.



\subsection{Rendering and Optimization}

\subsubsection{Rendering}
For rendering, we adopt the 2D Gaussian Splatting (2DGS) pipeline introduced in \cite{2dgs}. For further details on the underlying rendering mechanics, we refer readers to \cite{2dgs}.

\begin{figure*}[t]
  \centering
  \includegraphics[width=0.95\linewidth]{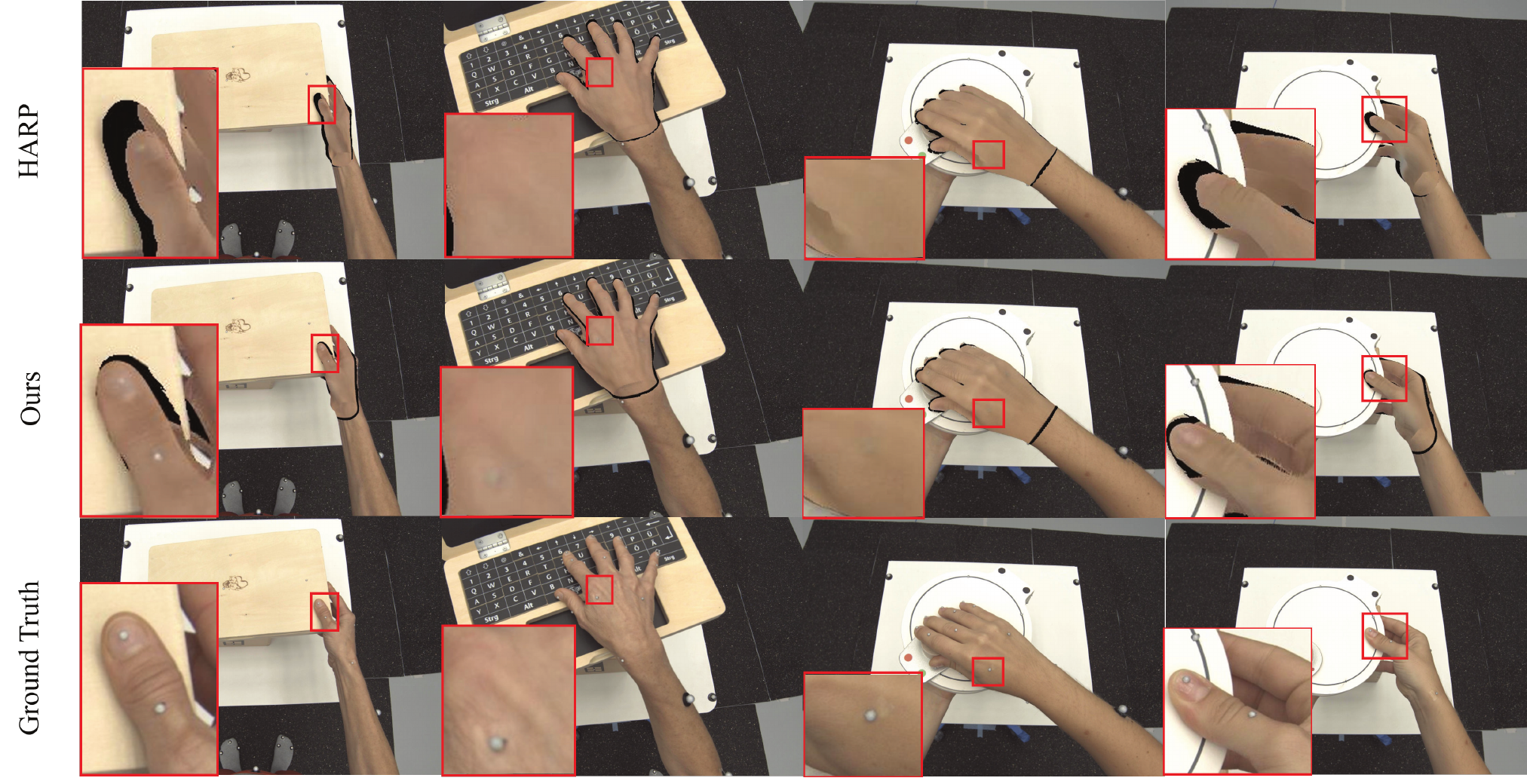}
  \caption{Comparison of rendering results with HARP on the ARCTIC dataset. Test results are shown for the sequences 's01 laptop use' and 's02 waffle use'. Our method reconstructs the markers on the hand more clearly."}
  \label{fig:s1laptop}
\end{figure*}

\subsubsection{Optimization}
Below, we describe the loss functions and the two-stage training strategy in the differentiable optimization pipeline. In the first stage, the focus is on optimizing the geometry attributes of both the attached and detached surfels, including the geometry attributes and the low-frequency harmonics coefficients, to provide a coarse description of the surface appearance.  In the second stage, the geometry attributes learned in the first stage are fixed, and the optimization enables learning of the opacity mask, which controls the density of detached Gaussian surfels and the high-frequency harmonics coefficients. This stage refines the texture details, enabling the model to capture high-frequency appearance features.

We employ a combination of loss functions to supervise the optimization process. For further details on the depth distortion loss and the normal consistency loss, we refer readers to \cite{2dgs}. The total loss is defined as:
\begin{equation}
    \mathcal{L} = \lambda_{d}\mathcal{L}_{d} + \lambda_{n}\mathcal{L}_{n} + \lambda_{c}\mathcal{L}_{c} + \lambda_{sil}\mathcal{L}_{sil} + \lambda_{b}\mathcal{L}_{b},
\end{equation}
where each term corresponds to a specific supervision objective:
\begin{enumerate}
    \item \textbf{Depth Distortion Loss $\mathcal{L}_{d}$: }This $\mathcal{L}_{1}$ loss minimizes the depth disparity of intersecting 2D surfels in the image space, ensuring a consistent depth representation. 
    \item \textbf{Normal Consistency Loss $\mathcal{L}_{n}$: }This term aligns the gradient of the depth map with the normal vectors of the 2D surfels, ensuring smooth surface transitions. 
    \item \textbf{Image Reconstruction Loss $\mathcal{L}_{c}$: }To ensure photometric consistency, we combine an $\mathcal{L}_{1}$-based reconstruction term with a D-SSIM term for perceptual quality \cite{10.1145/3592433}. 
    \item \textbf{Silhouette Loss $\mathcal{L}_{sil}$: }This loss enforces alignment between the rendered silhouette of the surfels and a ground truth silhouette mask \cite{silhouetteloss}. We generate the ground truth mask using SAM-HQ \cite{samhq}. 
    \item \textbf{Binding Loss $\mathcal{L}_{b}$: }To ensure that Detached Gaussian surfels remain close to the hand geometry, we introduce a binding loss. This loss penalizes discrepancies between the deformed surfels and the corresponding original mesh surface. A cutoff threshold $\delta$ is applied to prevent surfels from being too restricted to the original mesh geometry. For a surfel $i$ and its associated mesh face $j$, the binding loss is defined as: 
    \begin{equation}
    \mathcal{L}_{b}=\sum_{i} \sum_{j}\omega_{ij}  \max(d_{ij},\delta) (1- \mathbf{n}_{i}^{\mathrm{T}}\frac{ \mathbf{e1}_{j} \times  \mathbf{e2}_{j}}{\left| \mathbf{e1}_{j} \times  \mathbf{e2}_{j}\right|})
    \end{equation}
    where $\omega_{ij}$ indicates whether the surfel $i$ belongs to mesh face $j$ ($\omega_{ij} = 1$) or not ($\omega_{ij} = 0$), $d_{ij}$ represents the distance between the surfel centroid and the mesh face, $\mathbf{n}_{i}$ refers to the normal vector of the $i$-th surfel, and $\mathbf{e1}_{j}$, $\mathbf{e2}_{j}$ are the edges of mesh face $j$, used to compute the face normal.
    
\end{enumerate}

\section{Experiment}

\subsection{Experimental Setting}
\textbf{Datasets and Baselines}:
To validate the robustness of MASS under diverse egocentric conditions, we evaluate on three challenging benchmarks:
\begin{enumerate}
    \item ARCTIC for complex hand–object interactions in egocentric videos;
    \item Hand Appearance for fine-grained texture and appearance reconstruction;
    \item InterHand2.6M to assess the generalization of hand rendering under stereo settings.
\end{enumerate}
We evaluated our method with state-of-the-art methods on the stable camera captured monocular dataset: Hand Appearance \cite{karunratanakul2023harp}, compared to previous SOTA works \cite{JIANG2025111426, karunratanakul2023harp}. In terms of quantitative evaluation on the Hand Appearance dataset, we follow the setting of \cite{JIANG2025111426} to split the dataset and resize images. 
Although our model is not designed to handle multiview input, we evaluated our method in the default setting(without additional multiview camera calibration supervision) with previous SOTA methods \cite{chen2021s2hand,jiang2022selfrecon,10.1145/3544548.3581027,JIANG2025111426,jiang2023AMVUR,HTML,weng_humannerf_2022_cvpr} on the InterHand2.6M \cite{Moon_2020_ECCV_InterHand2.6M} 5fps validation set. We follow the HandAvatar\cite{10.1145/3544548.3581027} evaluation process, including choosing training data and validation data, center-cropping images, and resizing.
We also evaluated our model and HARP on ARCTIC from an egocentric view at 1400*1000 pixels. To ensure fair comparison. The MANO parameters with 3D poses for HARP optimization are also provided by WiLoR.
Our method operates efficiently on a single RTX 4090 GPU, with rendering time costing less than half that of HARP, which is significantly faster than implicit methods and competitive with real-time 3DGS approaches. For details on the dataset and experimental settings, please refer to the supplementary materials.

\begin{figure*}[ht]
  \centering
  \includegraphics[width=0.8\linewidth]{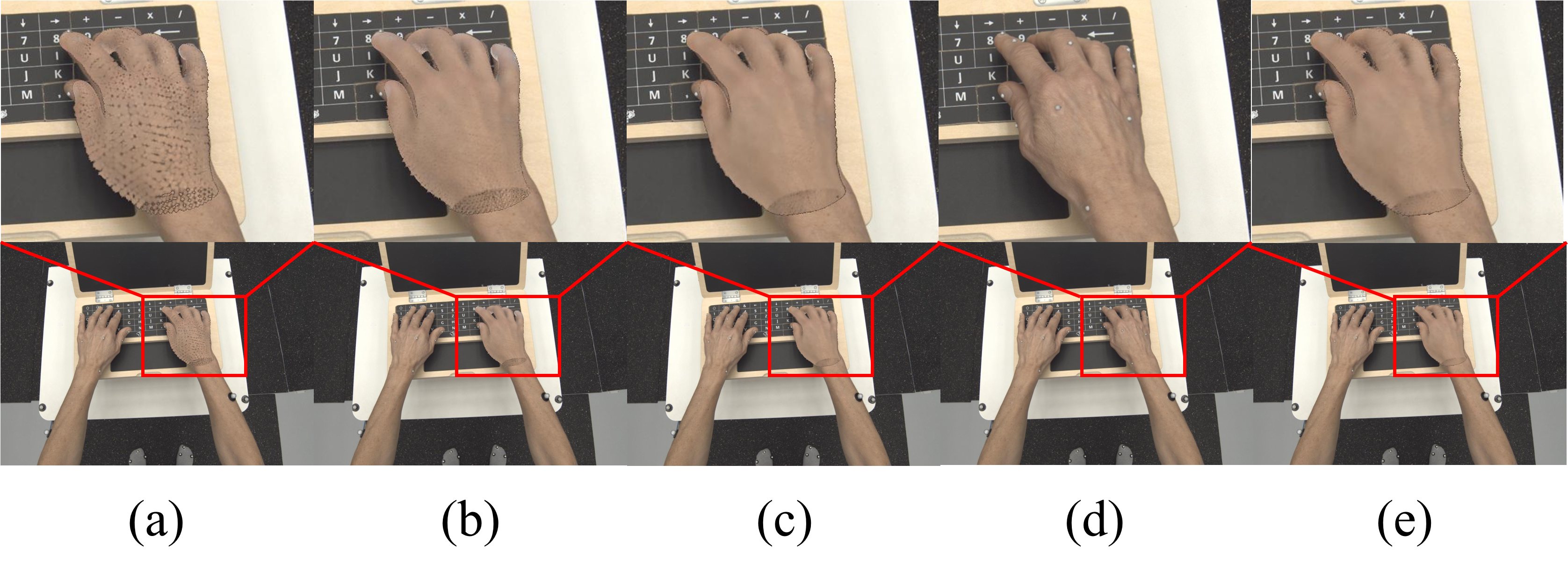}
  \caption{
  Qualitative evaluation of the ablation study on the ARCTIC dataset 
  \cite{fan2023arctic}. (a) method without Fractal Densification; (b) method without Detached Gaussian surfels; (c) full method; (d) referring to ground truth. (e) control setting with lower loop subdivision}
  \label{fig:ablation}
\end{figure*}

\begin{table}[t]
\centering
  \begin{tabular}{cccc}
  \toprule
     Methods & PSNR$\uparrow$  & LPIPS$\downarrow$  & SSIM $\uparrow$\\
  \midrule  
    HARP & 21.50 & 0.107  & 0.878\\
    3D-PSHR & 23.55 & 0.095 & 0.893\\
    Ours & $ \textbf{26.32}$ & $\textbf{0.091}$ & $\textbf{0.921}$\\
  \bottomrule
\end{tabular}
\caption{Quantitative evaluation on the Hand Appearance dataset \cite{karunratanakul2023harp}.}\label{tab:happ}
\end{table}

\begin{table}
\centering
\begin{tabular}{ccccc}
\toprule
\multirow{2}{*}{Methods}& \multirow{2}{*}{Views} & \multicolumn{3}{c}{val/Capture0} \\
 &  & LPIPS $\downarrow$ & PSNR $\uparrow$ & SSIM $\uparrow$ \\
\midrule
SelfRecon & 139 & 0.149 & 25.78 & 0.869 \\
HTML& 139 & 0.186 & 23.41 & 0.851 \\
$S^2$ HAND& 139 & 0.146 & 25.94 & 0.877 \\
AMVUR& 139 & 0.132 & 27.43 & 0.885 \\
HumanNeRF& 139 & 0.119 & 27.80 & 0.882 \\
HandAvatar& 139 & 0.106 & 28.04 & 0.890 \\
3D-PSHR& 139 & $\textbf{0.092}$ & 29.40 &  $\textbf{0.910}$ \\
Ours & 40 & 0.144 & $\textbf{30.28}$ & 0.905 \\
Ours & 20 & 0.177 & 27.78 & 0.887 \\
Ours & 10 & 0.198 & 27.20 & 0.860 \\
\bottomrule
\end{tabular}
\caption{Quantitive evaluation on Interhand 2.6M. Note that our method selected 40,20,10 training views rather than 139 views from the HandAvatar evaluation setting.}
\label{tab:interhands}
\end{table}

In addition, we test our model on the egocentric view of ARCTIC \cite{fan2023arctic}. The sequences are divided into $80\%$ for training and the rest for the test. We implement our method of rendering at 1400$\times$1000 resolution, compared with monocular-specified work HARP \cite{karunratanakul2023harp}.

\textbf{Implementation}: We implement our end-to-end training pipleine on the hand perception model WiLoR \cite{potamias2025wilorendtoend3dhand}, the segmentation model the ViT-H version SAM-HQ. And we built our deformation module on Dynamic-2DGS \cite{zhang2024dynamic2dgaussiansgeometrically}.
In terms of training setting, we set $\lambda_{c}=1,\lambda_{n}=0.02,\lambda_{d}=1000,\lambda_{sil}=1,\lambda_{b}=1$. For other implementation details, please refer to the supplementary materials.

\textbf{Metrics}: For rendering evaluation, we chose MS-SSIM, SSIM, PSNR, and LPIPS.


\begin{table}
\centering
  \begin{tabular}{ccccc}
    \toprule
    Methods & PSNR$\uparrow$  & LPIPS$\downarrow$  & MS-SSIM$\uparrow$ & Surfels\\
    \midrule
    full model  & 32.47 & 0.0223 & 0.984 & 4$\times$N \\
    model w/o $\mathcal{L}_{b}$ & 27.0 & 0.0612 & 0.953 & 4$\times$N\\
    w/o FD  & 30.60 & 0.0253 & 0.979 & N \\
    w/o DG & 31.20 & 0.0267& 0.977 & 4$\times$N\\ 
    LS$\downarrow$  & 32.20 & 0.0168& 0.986 & N \\ 
    \bottomrule
\end{tabular}
\caption{Quantitative evaluation of ablation study on “laptop” test set, N is 49216 surfels.}
\label{tab:freq}
\end{table}

\begin{figure}[t]
  \centering
\includegraphics[width=\columnwidth]{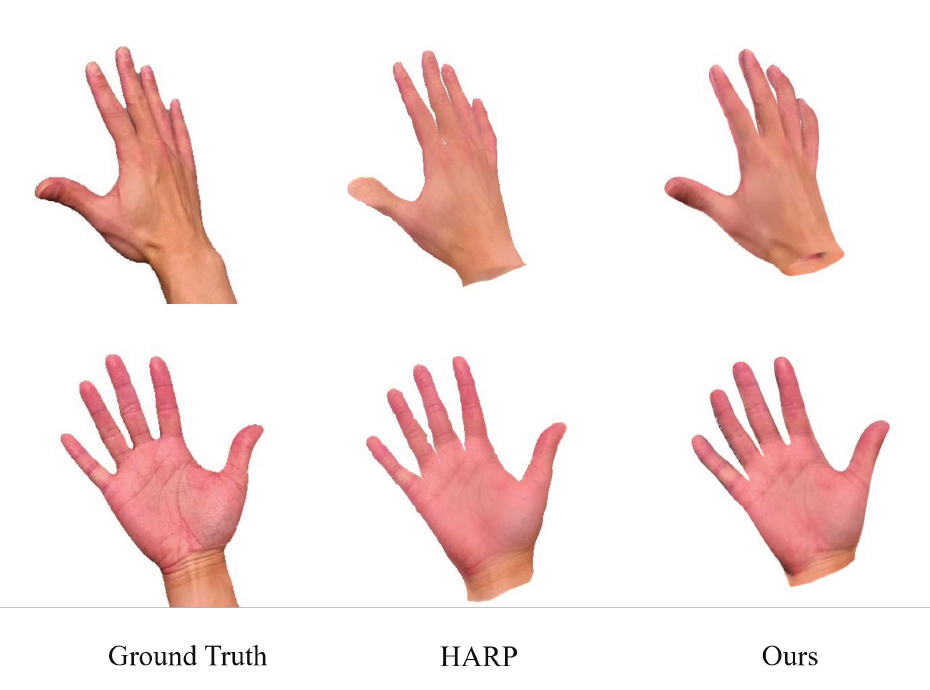}
  \caption{Visual comparison of the evaluation results of our method and the previous method on the hand appearance dataset \cite{karunratanakul2023harp}. Our rendering results have smoother boundaries and more detailed textures.}
  \label{fig:vishapp}
\end{figure}

\begin{figure}[t]
  \centering
\includegraphics[width=\columnwidth]{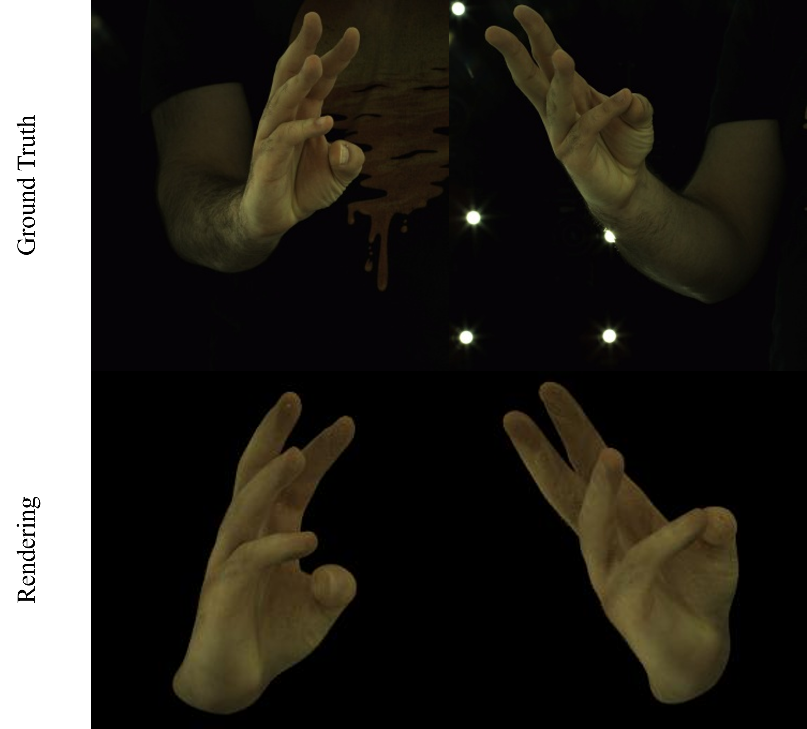}
  \caption{Visual result of our method on the Interhand 2.6M dataset \cite{Moon_2020_ECCV_InterHand2.6M}.}
  \label{fig:visinterhand}
\end{figure}

\subsection{Comparison with State-of-the-art Methods}
For the experiment comparing HARP on the ARCTIC  \cite{fan2023arctic}, we modified the code of HARP to cancel the arm modeling and data preparation part to unify the MANO pose and shape provided for fair comparison. In Quantitative evaluation, we outperform the HARP on all metrics, PSNR, MS-SSIM, and LPIPS in Table \ref{tab:arctic_comps} except the LPIPS metric in the capsule case, where our LPIPS is slightly higher. Additional experiment results on different subjects and interacting objects are shown in Fig.\ref{fig:histogram}. For comparison visualization shown in Fig.\ref{fig:s1laptop}, our method yields results with less misalignment and reconstructs the detailed appearance like the markers on the hand. 

\begin{figure}[t]
  \centering
\includegraphics[width=\columnwidth]{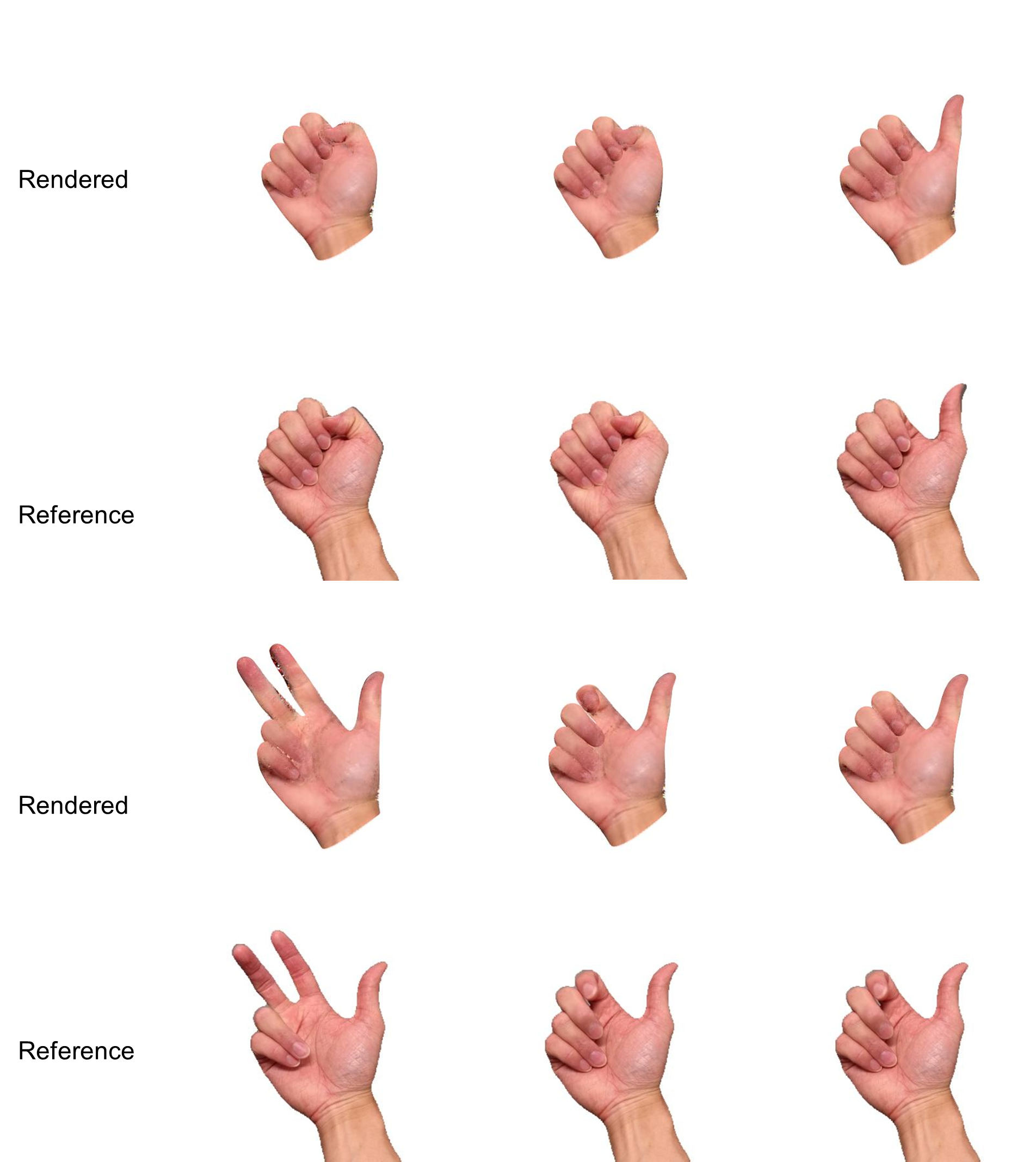}
  \caption{Rendering result of the experiment on Hand Appearance. For poses such as a closed fist, reference images show subtle shadows between fingers, but our model fails to reconstruct this effect. }
  \label{fig:vishappfail}
\end{figure}

We compared our MASS model with HARP \cite{karunratanakul2023harp} and 3D-PSHR \cite{JIANG2025111426} on Hand Appearance. HARP and 3D-PSHR fit the hand with arm modeling. For fair comparison, we compute the evaluation metrics in the provided masked area. The quantitative result of the experiment on the Hand Appearance dataset is shown in Table \ref{tab:happ}. Our model performs better than HARP on PSNR, LPIPS, and SSIM. For visualization comparison, we compared with HARP and canceled arm modeling. The result is shown in Fig.\ref{fig:vishapp}. 
We compared inference efficiency metrics in Table \ref{tab:realtime}, evaluating the real-time performance. Additionally, our method trains on a single ARCTIC subsequence in 5.5 minutes, speedup over HARP's 50 minutes, while achieving higher fidelity. 
While our method captures fine details like skin texture and markers (Fig.\ref{fig:s1laptop}), it struggles with inter-finger occlusions. As shown in the Fig.\ref{fig:vishappfail}, when fingers are tightly pressed together, the model fails to reconstruct the subtle shadowing between them. It is a challenge inherent to monocular methods lacking explicit lighting or global shading priors.

In addition, we tested our method on Interhand2.6M 5fps validation capture. The quantitative result is reported in Table \ref{tab:interhands}. Our method provides a competitive performance with fewer training views. For other subjective analyses and visual results, please refer to the supplementary materials.

\subsection{Ablation Study}
We evaluated the effectiveness of the surfels Fractal Densification, Detached Gaussian surfel Deformation, and Binding Loss. Quantitative result is shown in Table \ref{tab:freq}.  From Fig.\ref{fig:ablation}, we observe that the render result is smooth from the model without detached Gaussian surfels. The mesh-aligned surfels reconstruct the template-aligned surface, relying heavily on the accuracy of the provided pose. As for the surfels without Fractal Densification, we notice that the 2D surfels did not cover the area of the wrist. 2D Gaussian surfels from the MANO template without densification are insufficient, resulting in a mass of artifacts. Note that the training process without the surface binding loss did not converge, so we did not offer a visualized result. 
To properly evaluate the contribution of Fractal Densification (FD), we conducted an additional controlled ablation study where we reduced the subdivision level of the mesh to ensure an identical total number of Gaussians between the full method and the variant without Fractal Densification. This creates a fair comparison where the only variable is the initialization pattern. 

\begin{figure}[t]
  \centering
  \includegraphics[width=0.8\linewidth]{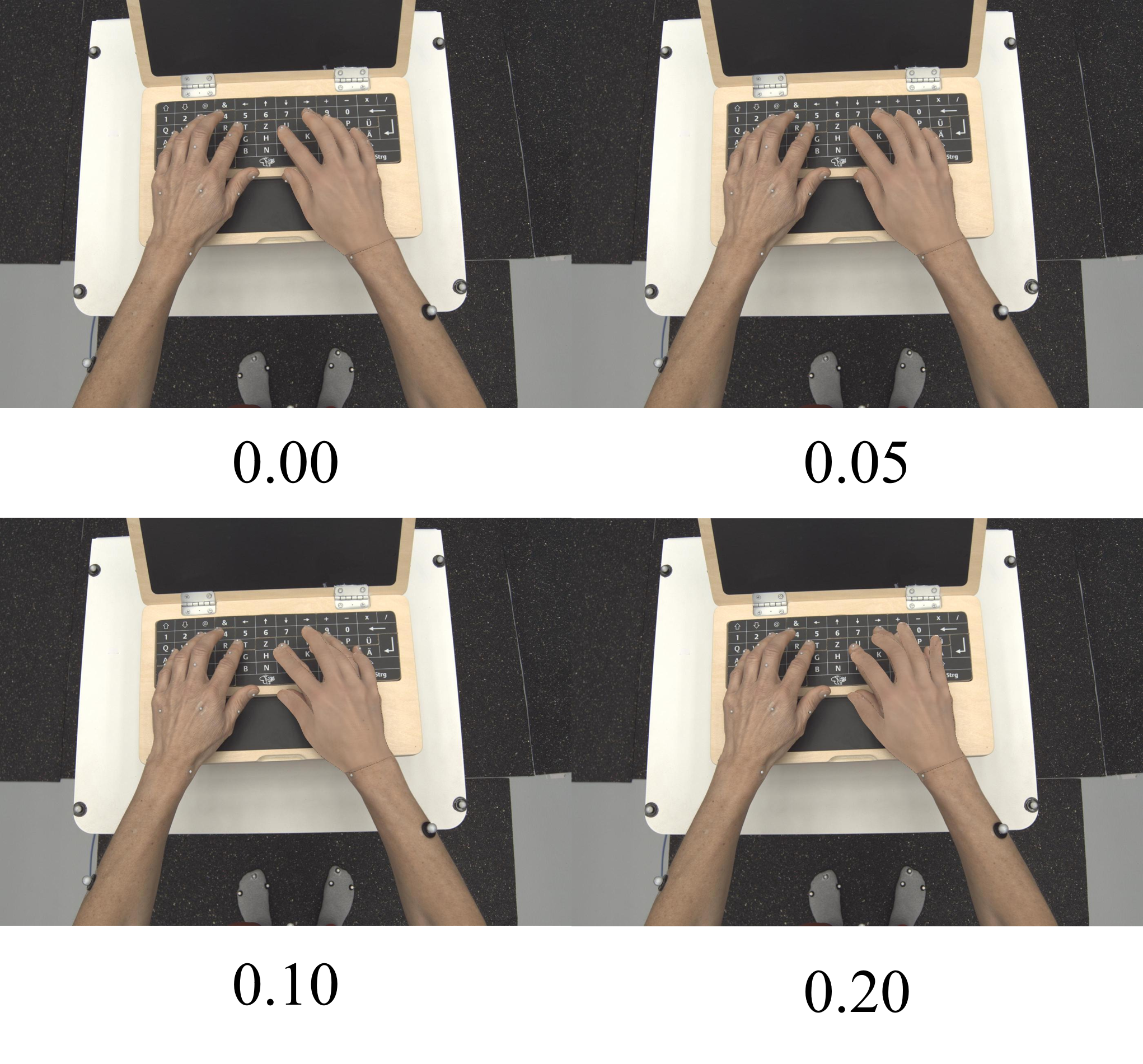}
  \caption{
  Qualitative evaluation of the noise robustness ablation study on ARCTIC dataset.}
  \label{fig:noise}
\end{figure}

As shown in Table \ref{tab:freq}, the full model with Fractal Densification achieves a PSNR of 32.47. The model with Fractal Densification and lower loop subdivision achieves 32.20 PSNR, while the variant without FD achieves 30.60 PSNR. This 1.6 dB improvement demonstrates that FD provides significant benefits beyond simply increasing the number of Gaussians. The fractal geometry pattern specifically targets high-frequency regions near vertices,  resulting in a more accurate representation of fine hand details.
\begin{table}
\centering
  \begin{tabular}{ccccc}
    \toprule
      & HandAvatar & HARP & 3D-PSHR & Ours\\
    \midrule
    FPS & 0.3 & 26 & - & 58 \\
    \bottomrule
\end{tabular}
\caption{
Real-time performance in inference on the Interhand2.6M.}
\label{tab:realtime}
\end{table}
To evaluate the robustness of our method to tracker failures, we analyzed system behavior under noisy pose estimates. Specifically, we injected Gaussian noise into the MANO pose parameters generated by the WiLoR tracker to simulate real-world tracking errors. Based on the results in Table \ref{tab:noise}, our method demonstrates robustness to pose noise. With 0.2 standard deviation joint-angle noise, our model exhibits only a 0.71 dB PSNR degradation. However, from the inference visualization Fig. \ref{fig:noise}, we can still find the noise impact on the rendering.

\begin{table}
\centering
  \begin{tabular}{cccc}
    \toprule
    PSNR$\uparrow$  & LPIPS$\downarrow$  & MS-SSIM$\uparrow$ & Noise Std\\
    \midrule
    31.77 & 0.0242 & 0.982 & 0 \\
    31.58 & 0.0244 & 0.982 & 0.05 \\
    31.42 & 0.0244 & 0.981 & 0.1 \\
    31.06 & 0.0246 & 0.980 & 0.2 \\ 
    \bottomrule
\end{tabular}
\caption{
Quantitative evaluation of ablation study on “laptop” test set, noise level is indicated by standard deviation.}
\label{tab:noise}
\end{table}

We conduct an ablation study to validate the effectiveness of three key components in MASS: Fractal Densification (FD), Detached Gaussian Surfel Deformation (DG), and the Binding Loss. Quantitative results on the “laptop” sequence from ARCTIC are reported in Table \ref{tab:freq}, with qualitative comparisons in Fig.\ref{fig:ablation}.
\begin{enumerate}
    \item Fractal Densification (FD) ensures high surfel coverage, especially in high-curvature or under-sampled regions such as the wrist and finger joints. Without FD Figure \ref{fig:ablation} (a), surfels fail to span the full hand surface, leading to visible holes and reconstruction artifacts—particularly where the base MANO mesh is coarse.
    \item Detached Gaussian Surfel Deformation (DG) enables fine-grained geometric adaptation beyond the constraints of the parametric mesh. When DG is disabled (Fig.\ref{fig:ablation} (b)), the model reverts to mesh-aligned surfels only, producing overly smooth results that cannot capture personalized details like skin wrinkles or motion-induced deformations.
    \item Binding Loss plays a critical role in stabilizing early-stage optimization by softly anchoring detached surfels to their source mesh faces. Training without $\mathcal{L}_b$ leads to severe instability and divergence. The quantitative result is not meaningful, as the scale of surfels enlarges uncontrollably from the hand surface and covers the whole viewing space.
    \item 
    Loop Subdivision (LS) refines the low poly of MANO by registering 2d surfels on the subdivided mesh faces. We set up this controlled experiment to compare the impact of geometry pattern in initialization.
\end{enumerate}

Together, these components enable MASS to achieve both geometric fidelity and deformation flexibility, striking a balance between structure-aware initialization and data-driven refinement.

\section{Conclusion and Discussion}

\subsection{Real-World Generalization and Limitations}
\begin{figure}[t]
  \centering
\includegraphics[width=\columnwidth]{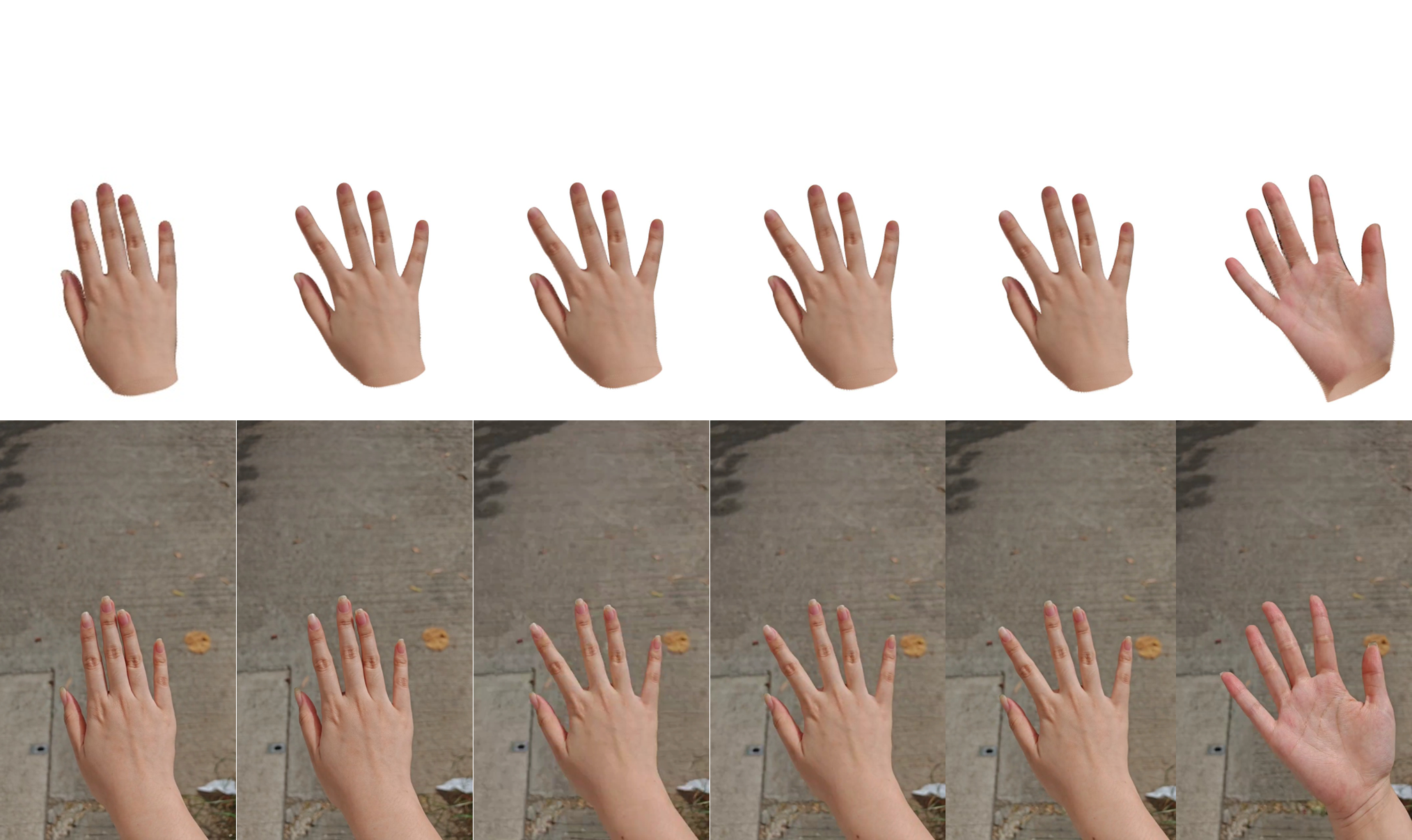}
  \caption{Rendering and reconstruction on the in-the-wild video.}
  \label{fig:visinthewild1}
\end{figure}

\begin{figure}[t]
  \centering
\includegraphics[width=\columnwidth]{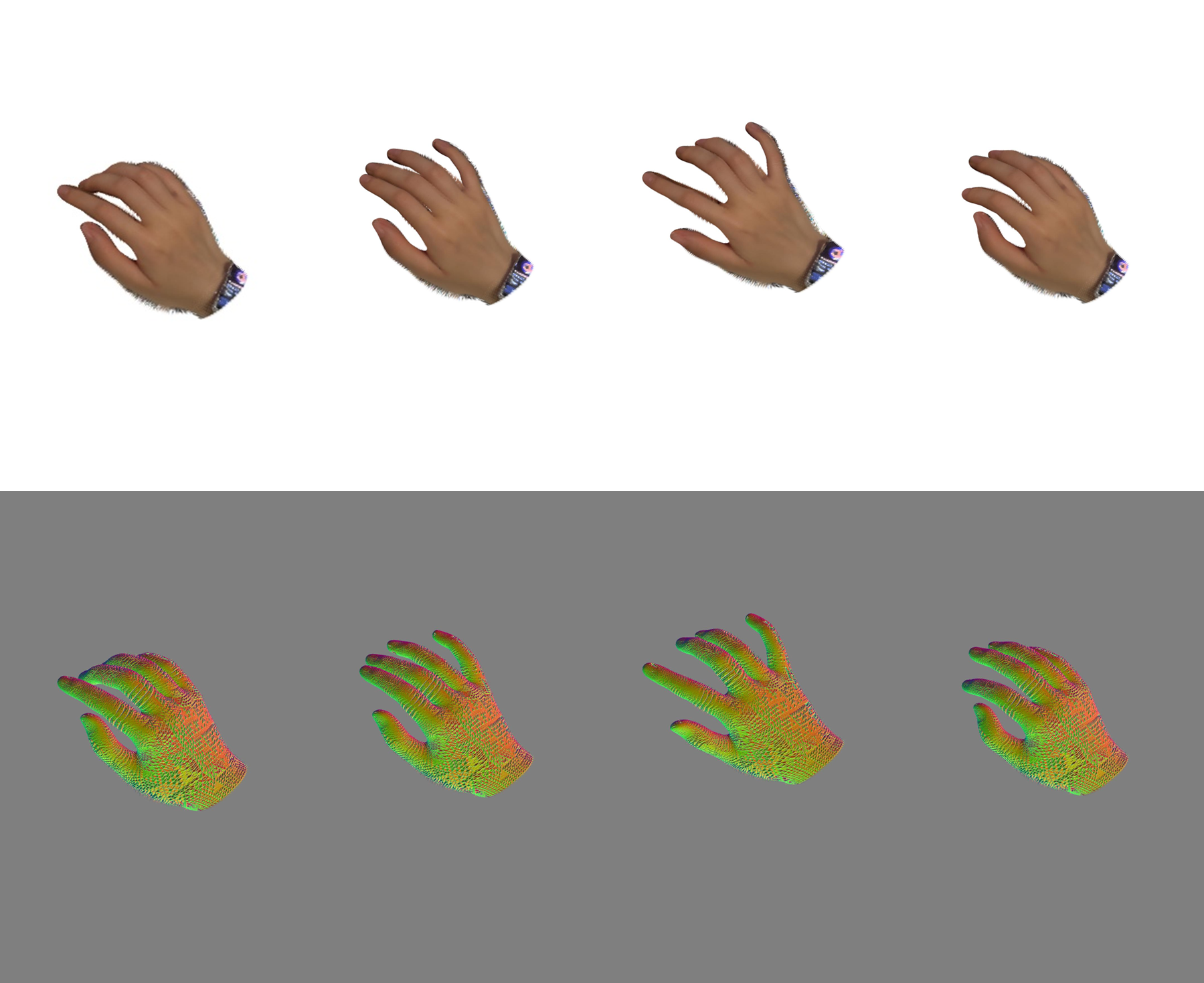}
  \caption{Rendering and reconstruction on the in-the-wild video with accessories. }
  \label{fig:visinthewild2}
\end{figure}

Beyond controlled benchmarks, we evaluate MASS on in-the-wild egocentric videos captured with consumer phones (Fig.\ref{fig:visinthewild1}). Our method successfully reconstructs hands under diverse lighting, textures, and interactions (typing, mouse use). Notably, it can even render accessories like smartwatches (Fig.\ref{fig:visinthewild2}) though their geometry is not reconstructed due to a lack of multiview supervision or explicit object modeling. This highlights a key limitation: while MASS excels at appearance capture, it currently models only the hand surface, not its interaction with surrounding objects or dynamic occlusions. Failures often stem from inaccurate segmentation masks or reflections that confuse the silhouette loss, suggesting future work should incorporate scene context or joint object-hand modeling.


\subsection{Conclusion}
We have presented Mesh-inellipse Aligned deformable Surfel Splatting (MASS), a novel framework for high-fidelity 3D hand reconstruction from egocentric monocular video. By bridging parametric hand priors with 2D Gaussian surfel splatting, MASS achieves unprecedented geometric detail and rendering quality under challenging monocular and egocentric conditions as state-of-the-art methods in both quantitative metrics and visual fidelity.

A key insight of our work is that surface-aligned surfel initialization via the Steiner Inellipse provides a geometrically principled foundation for stable optimization, especially when depth cues are weak. While MASS is instantiated for hand reconstruction, we hypothesize that the core paradigm—mesh-aligned surfel initialization via geometric primitives followed by neural deformation with hashgrid-MLP refinement—is broadly applicable to other articulated or deformable objects. In settings where a coarse mesh or depth prior is available (\eg, from LiDAR, stereo, or monocular depth estimation), our surfel conversion pipeline could serve as a lightweight, geometry-aware scaffold for high-fidelity Gaussian splatting at the object or scene level. 

MASS has limitations. Because our appearance model uses view-dependent spherical harmonics without explicit lighting disentanglement, reconstructions are sensitive to strong or varying illumination—an inherent challenge in monocular settings. Additionally, by operating in camera coordinates, our method sidesteps the need for accurate camera trajectory estimation but sacrifices world-scale consistency, limiting applications that require absolute 3D positioning.

Looking ahead, we hypothesize that the mesh-to-surfel conversion paradigm introduced here could generalize beyond hands—to faces, full bodies, or even articulated robotic systems—where a coarse template must be enriched with fine, data-driven geometry. Integrating neural reflectance models or differentiable shaders could further enable editable relighting and material editing. In summary, MASS demonstrates that structured geometric initialization, when combined with efficient neural deformation, offers a promising path toward real-world, real-time avatars from casually captured video.

\section*{Acknowledgement}

The research work described in this paper was conducted in the JC STEM Lab of Machine Learning and 
Computer Vision funded by The Hong Kong Jockey Club Charities Trust. This research received partially 
support from the Global STEM Professorship Scheme from the Hong Kong Special Administrative Region.

{\small
\bibliographystyle{cvm}
\bibliography{main}
}

\end{document}